\documentclass[11pt]{article}
\usepackage[letterpaper, margin=1in]{geometry}
\usepackage{microtype}
\usepackage{hyperref}
\usepackage{url}
\usepackage{natbib}
\usepackage{booktabs}
\usepackage{graphicx}
\usepackage{amsmath,amssymb,amsthm}
\usepackage{tikz}
\usetikzlibrary{arrows.meta, decorations.pathreplacing, calc, positioning, fadings, backgrounds}
\definecolor{darkblue}{rgb}{0, 0, 0.5}
\definecolor{correctgreen}{HTML}{2E8B57}
\definecolor{incorrectred}{HTML}{C04040}
\definecolor{querygray}{HTML}{555555}
\definecolor{anglecolor}{HTML}{1F4E8C}
\hypersetup{colorlinks=true, citecolor=darkblue, linkcolor=darkblue, urlcolor=darkblue}
\newtheorem{definition}{Definition}
\title{How Transformers Reject Wrong Answers:\\
       Rotational Dynamics of Factual Constraint Processing}
\author{Javier Mar\'in\thanks{\texttt{javier@groundlens.dev}} \\[2pt]
        Groundlens-dev}
\date{\small Revised version, \today\ (supersedes arXiv:2603.13259v1)}

\begin{document}
\maketitle
\begin{abstract}
When a decoder-only transformer is forced to process matched correct and incorrect single-token continuations of a factual query, the two pathways through hidden-state space diverge in a specific way: displacement vectors from the query-only representation maintain approximately equal magnitude but rotate apart in direction. The angular separation grows through mid-depth, and late layers resolve the asymmetric outcome---a logit-lens preference that, in the incorrect run, falls far below the naive prior of equal probability, corresponding to the model assigning approximately $11.5\times$ more probability to the incorrect token than to the correct one. We characterize this two-phase pattern---rotational divergence in mid-depth followed by late-layer asymmetric commitment---as the empirical geometric signature of what looks externally like the model ``rejecting'' a wrong continuation, while remaining explicit that it is an \emph{observational} characterization, not a causal account, and that the incorrect-run trajectory admits a second reading (the model conforming to the token it is forced to carry) that only a random-token control can separate. The pattern is consistent across six decoder-only transformers with measurable factual processing, spanning four architecture families (Llama, Mistral, Gemma, StableLM) from 1B to 13B parameters; a seventh model (Qwen2 1.5B), the fifth family, shows a flat profile under the present extraction protocol that is plausibly a tokenizer-fragmentation artefact rather than a real scale floor, so the question of an emergence threshold is left open. Linear probes recover the correct/incorrect distinction at intermediate depth in all six measurable models, and cross-domain probe transfer is structurally asymmetric---a financial--medical corridor transfers far better than transport pairs---a regularity that replicates across all six architectures. On the two models where single-layer activation patching is cleanly interpretable (LLaMA-2 13B and Mistral 7B), patching from the correct run into the incorrect run does not produce a layer band with consistent recovery of the correct token; a third patched model (StableLM-2 1.6B) recovers uniformly at every layer including the first and above the recovery ceiling, a pattern we diagnose as an artefact of its patching code path and exclude. Under this scoped result the late-layer asymmetry is not localized to a single discrete component in the regime tested. Taken together, the evidence is consistent with a distributed-by-trajectory account of factual constraint processing---geometric structure that emerges cumulatively across many layers rather than from a single localized circuit---and inconsistent with the simplest single-layer localized-recall account. We document this structure with controlled forced-completion probing across seven models, three professional domains, and 300 stratified queries.
\end{abstract}
\section{Introduction}\label{sec:intro}
Transformer language models generate text that can be factually incorrect. Such failures fall under the broader heading of factual hallucination, a term whose definition varies considerably across the literature; we adopt a deliberately narrow operational view of factual correctness, detailed in Section~\ref{sec:related}. Understanding the internal processing that differentiates correct from incorrect factual continuations is a prerequisite for principled mitigation. A growing body of work has established that hidden representations encode linearly separable features related to truthfulness \citep{burns2023discovering, azaria2023internal, li2024inference, marks2024geometry}, that factual associations can be localized to specific layers and components \citep{meng2022locating, geva2023dissecting, geva2021transformer}, and that intermediate representations can be decoded into output distributions via the logit lens \citep{nostalgebraist2020logitlens, belrose2023eliciting}. These approaches are mainly static: a probe at a particular depth, a feature direction in a single layer, an intervention at a localized component. Less is known about the layerwise dynamics: how the distinction between correct and incorrect factual continuations emerges, evolves, and resolves across the full depth of the network.
We address this gap with a controlled experimental instrument and a layerwise geometric analysis. Our method, forced-completion probing, builds queries with exactly one single-token correct continuation and exactly one single-token incorrect continuation, and measures five complementary quantities at every layer for both forced runs: trajectory similarity, displacement geometry, linear probe accuracy, logit-lens commitment, and attention allocation. By forcing the model to process both completions of the same query, we obtain matched pairs that isolate the effect of factual correctness from confounds such as token frequency or syntactic complexity. The protocol is methodologically conservative in the following sense: it observes geometric structure under controlled forced continuations but does not by itself bridge to the geometry of hallucinations produced under unconstrained autoregressive generation. We are explicit about this scope throughout.
\paragraph{Scope and stance.} We document the layerwise geometric structure that emerges in this regime, characterize its scale and architecture dependence, and report a null result on single-layer activation patching that bounds the causal account our data can support. The null is scoped: it holds on the two architectures where the patch is cleanly interpretable (LLaMA-2 13B and Mistral 7B), while a third patched model (StableLM-2 1.6B) produces a profile we diagnose as an implementation artefact and exclude (Section~\ref{sec:patching}, Appendix~\ref{app:patching}). We are deliberately conservative on interpretation: the data are consistent with a distributed-by-trajectory account in which the geometric structure is the cumulative consequence of many small contributions across the network, and also consistent in principle with a localized mechanism that single-layer patching fails to detect. We do not adjudicate between these accounts; we document the structure, report the null result and the excluded artefact plainly, and frame the question of mechanism as open.
\section{Related Work}\label{sec:related}
\paragraph{Probing factual representations and recall mechanisms.} A growing line of work has documented that hidden representations encode linearly separable features related to truthfulness \citep{burns2023discovering, azaria2023internal, li2024inference, marks2024geometry}, that factual associations can be localized to mid-layer MLP modules \citep{meng2022locating, geva2023dissecting, geva2021transformer, dai2022knowledge}, and that knowledge can be edited at those localized sites. These analyses are predominantly static (a probe at a particular depth, an intervention at a localized component) and focus on simple fact-completion tasks. We complement them with a dynamic, layer-by-layer account of matched correct vs.\ incorrect forced continuations, and we treat our null result on single-layer patching (Section~\ref{sec:patching}) as evidence that the localization properties documented in this line do not transfer cleanly to the forced-continuation regime, on the architectures where our patch is cleanly interpretable.
\paragraph{Representation geometry and intermediate-layer decoding.} The linear representation hypothesis \citep{park2024linear} and work on structured representations \citep{gurnee2024language, dar2023analyzing, hernandez2024linearity, ethayarajh2019contextual} establish that transformer hidden states carry geometric structure. The logit lens provides a way to decode intermediate representations through the unembedding projection \citep{nostalgebraist2020logitlens, belrose2023eliciting, din2023jump}. Our approach combines both: a controlled protocol that decomposes displacement geometry into radial and angular components and that uses the logit lens to define commitment ratios for the model's intermediate preference between correct and incorrect tokens.
\paragraph{Scope within the hallucination literature.} A model produces a hallucination when its output is not grounded in the facts it should reflect. The literature slices this differently: \citet{maynez2020faithfulness} separates faithfulness from factuality; \citet{ji2023survey} and \citet{huang2023survey} catalog the broader taxonomy; \citet{tam2023evaluating} measures factual consistency in generated summaries; \citet{liu2026world} reframes the question in terms of the model's internal world-model. Across these framings, factually incorrect token-level continuations are recognised as one form of hallucination.
This paper does not measure free-generation hallucination. It measures the hidden-state geometry produced when a model is conditioned to output a specific factually-correct or factually-incorrect single-token continuation. The relevance to the broader hallucination question is direct: the geometric signature we document---rotational divergence of equal-magnitude displacement vectors---appears under conditioning, before the model commits to a token. Whether the same signature appears under free generation, and whether it can be used as an early-warning indicator, is the natural next experiment and is beyond the present scope (Section~\ref{sec:limitations}).
\section{Methods}\label{sec:methods}
\subsection{Forced-Completion Probing}\label{sec:fcp}
We define forced-completion probing as the following protocol (Figure~\ref{fig:method}). For each query $q$ we run three forward passes: a \emph{correct run} in which the correct single-token continuation $t^{+}$ is appended to the query, an \emph{incorrect run} in which an incorrect but domain-meaningful single-token continuation $t^{-}$ is appended, and a \emph{query-only pass} with no token appended as a baseline. At every layer $\ell$ we extract the residual-stream hidden state at the response-token position, yielding $h_\ell^{+}, h_\ell^{-} \in \mathbb{R}^d$ from the completion runs and $h_\ell^{q}$ from the query-only pass. Hidden states (\texttt{float16}, per the model load in Section~\ref{sec:models}) are upcast to \texttt{float32} before geometric computation.
\begin{figure}[h]
\centering
\resizebox{\textwidth}{!}{%
\begin{tikzpicture}[
  >={Latex[length=2.0mm, width=1.4mm]},
  font=\small\sffamily,
]
\begin{scope}[shift={(0, 0)}]
  \node[draw=querygray, rounded corners=2pt, fill=querygray!8,
        font=\small\sffamily, align=center, inner sep=4pt,
        minimum width=3.4cm, minimum height=0.9cm] (qA)
    at (0, 1.7)
    {query \(q\) \\[1pt] {\footnotesize\itshape ``The capital of France is''}};
  \coordinate (splitA) at (0, 0.7);
  \draw[querygray, line width=0.4mm] (qA.south) -- (splitA);
  \draw[->, correctgreen, line width=0.5mm]
    (splitA) -- ++(-1.2, -0.5) coordinate (cposEnd);
  \node[draw=correctgreen, rounded corners=2pt, fill=correctgreen!10,
        font=\footnotesize, align=center, inner sep=3pt,
        minimum width=1.7cm, minimum height=0.8cm]
    at ($(cposEnd) + (0, -0.45)$)
    {\(t^{+}\) \, (correct) \\[1pt] ``Paris''};
  \draw[->, incorrectred, line width=0.5mm]
    (splitA) -- ++(1.2, -0.5) coordinate (cnegEnd);
  \node[draw=incorrectred, rounded corners=2pt, fill=incorrectred!10,
        font=\footnotesize, align=center, inner sep=3pt,
        minimum width=1.7cm, minimum height=0.8cm]
    at ($(cnegEnd) + (0, -0.45)$)
    {\(t^{-}\) \, (incorrect) \\[1pt] ``London''};
  \node[font=\sffamily\small\bfseries, querygray]
    at (0, -2.0) {(a) \, Matched pair};
\end{scope}
\begin{scope}[shift={(5.6, 0)}]
  \draw[draw=querygray, fill=querygray!5, rounded corners=2pt]
    (-0.85, -0.7) rectangle (0.85, 1.8);
  \foreach \y in {-0.3, 0.1, 0.5, 0.9, 1.3} {
    \draw[querygray!30, very thin] (-0.85, \y) -- (0.85, \y);
  }
  \node[font=\footnotesize, querygray] at (0, -0.95) {\(L\) layers};
  \draw[->, querygray, line width=0.4mm] (-2.0, 1.4) -- (-0.85, 1.4);
  \node[font=\footnotesize, anchor=east, querygray] at (-2.0, 1.4) {\(q\)};
  \draw[->, correctgreen, line width=0.5mm] (-2.0, 0.55) -- (-0.85, 0.55);
  \node[font=\footnotesize, anchor=east, correctgreen] at (-2.0, 0.55) {\(q + t^{+}\)};
  \draw[->, incorrectred, line width=0.5mm] (-2.0, -0.3) -- (-0.85, -0.3);
  \node[font=\footnotesize, anchor=east, incorrectred] at (-2.0, -0.3) {\(q + t^{-}\)};
  \draw[->, querygray, line width=0.4mm] (0.85, 0.55) -- ++(1.1, 0);
  \node[font=\footnotesize, anchor=west, querygray, align=left]
    at (1.95, 0.55) {hidden states \\ \(h_\ell\) per layer};
  \node[font=\sffamily\small\bfseries, querygray]
    at (0, -2.0) {(b) \, Three forward runs};
\end{scope}
\begin{scope}[shift={(11.0, 0.3)}]
  \filldraw[querygray] (0, 0) circle (1.4pt);
  \node[anchor=north east, font=\footnotesize, querygray] at (0.05, -0.05) {\(h_\ell^{q}\)};
  \def\angA{60}
  \def\angB{120}
  \def\magn{1.7}
  \draw[->, correctgreen, line width=0.65mm]
    (0, 0) -- ({\magn*cos(\angA)}, {\magn*sin(\angA)}) coordinate (vplus);
  \node[anchor=west, font=\small\bfseries, correctgreen]
    at ($(vplus) + (0.05, 0)$) {\(d_{\ell}^{+}\)};
  \draw[->, incorrectred, line width=0.65mm]
    (0, 0) -- ({\magn*cos(\angB)}, {\magn*sin(\angB)}) coordinate (vminus);
  \node[anchor=east, font=\small\bfseries, incorrectred]
    at ($(vminus) + (-0.05, 0)$) {\(d_{\ell}^{-}\)};
  \draw[anglecolor, line width=0.35mm]
    ({0.55*cos(\angA)}, {0.55*sin(\angA)})
    arc[start angle=\angA, end angle=\angB, radius=0.55];
  \node[anglecolor, font=\small] at (0, 0.80) {\(\theta(\ell)\)};
  \node[font=\footnotesize, querygray, align=center, text width=4.0cm]
    at (0, -0.95)
    {\(\|d_\ell^{+}\| \approx \|d_\ell^{-}\|\), \\ \(\cos\theta(\ell)\) decreases with depth};
  \node[font=\sffamily\small\bfseries, querygray]
    at (0, -2.3) {(c) \, Geometric divergence};
\end{scope}
\end{tikzpicture}%
}
\caption{\textbf{Forced-completion probing.} A matched single-token pair (a) is processed through three forward passes (b), yielding the layerwise displacement geometry that is the object of study (c). Notation and formal definitions are given in Section~\ref{sec:fcp}.}
\label{fig:method}
\end{figure}
The protocol's strength is that the matched-pair design isolates the effect of factual correctness from token-frequency, syntactic-complexity and length confounds. Its limit is that a forced continuation is not free generation, and that the incorrect-run trajectory does not by itself distinguish rejection of a wrong fact from accommodation of the token the model is compelled to carry; the random-token control of Section~\ref{sec:limitations} is designed to separate the two.
\subsection{Dataset Design}\label{sec:dataset}
We manually generated 300 queries stratified across three domains (financial compliance, medical protocols, transport regulation) with $n=100$ each and three category types: deep constraint ($n=182$, correct answer requires integrating a domain-specific factual rule with the query context where a surface cue would favor the incorrect answer); control ($n=66$, correct answer is determinate and surface and deep cues agree); and neutral ($n=52$, the correct/incorrect distinction does not depend on domain-specific factual reasoning, serving as a within-protocol null control following \citet{hewitt2019designing}). Each query is human-authored such that (i) the correct $t^+$ and incorrect $t^-$ continuations both map to single BPE tokens under the studied tokenizers, (ii) both are domain-meaningful rather than arbitrary distractors, and (iii) deep-keyword and surface-keyword spans are character-offset-aligned to the query string for attention analysis. The complete per-domain category mapping, sample queries per category, and full annotation protocol are in Appendix~\ref{app:dataset}.
\subsection{Models}\label{sec:models}
We evaluate seven decoder-only transformer models across five architecture families: LLaMA-2 13B \citep{touvron2023llama}, Mistral 7B v0.3 \citep{jiang2023mistral}, Llama 3.2 3B \citep{dubey2024llama}, Gemma 2 2B \citep{gemmateam2024}, StableLM-2 1.6B \citep{bellagente2024stable}, Llama 3.2 1B \citep{dubey2024llama}, and Qwen2 1.5B \citep{yang2024qwen2}. Model parameters and layer counts are in Table~\ref{tab:models}. All models are loaded in \texttt{float16} with \texttt{attn\_implementation="eager"}.
\begin{table}[h]
\centering
\caption{Decoder-only transformer models evaluated.}
\label{tab:models}
\small
\begin{tabular}{@{}lrrr@{}}
\toprule
Model & Params & Layers & $d$ \\
\midrule
LLaMA-2 13B       & 13B   & 40 & 5120 \\
Mistral 7B v0.3   & 7B    & 32 & 4096 \\
Llama 3.2 3B      & 3B    & 28 & 3072 \\
Gemma 2 2B        & 2B    & 26 & 2304 \\
StableLM-2 1.6B   & 1.6B  & 24 & 2048 \\
Llama 3.2 1B      & 1B    & 16 & 2048 \\
Qwen2 1.5B        & 1.5B  & 28 & 1536 \\
\bottomrule
\end{tabular}
\end{table}
\subsection{Measurements}\label{sec:measurements}
We extract five complementary measurements per query, per layer, per run.
\begin{itemize}
    \item \textbf{(M1) Trajectory similarity}: $\tau(\ell) = \cos(h_\ell^+, h_\ell^-)$ between correct- and incorrect-run hidden states.
    \item \textbf{(M2) Displacement geometry}: displacement vectors $d_\ell^\pm = h_\ell^\pm - h_\ell^q$ decomposed into cosine similarity $\cos(d_\ell^+, d_\ell^-)$ and per-query norm ratio $\eta_i(\ell) = \|d_{i,\ell}^+\| / \|d_{i,\ell}^-\|$ (with mean $\bar{\eta}(\ell)$ across queries reported in Table~\ref{tab:displacement}), separating angular from radial divergence.
    \item \textbf{(M3) Linear probing}: logistic regression with 5-fold stratified cross-validation on hidden states at each layer to predict correct vs.\ incorrect run.
    \item \textbf{(M4) Logit-lens commitment ratio}: $\kappa(\ell) = \exp(\phi(h_\ell)[t^+]) / (\exp(\phi(h_\ell)[t^+]) + \exp(\phi(h_\ell)[t^-]))$, where $\phi: \mathbb{R}^d \to \mathbb{R}^{|V|}$ is the unembedding projection, using the normalized logit lens of \citet{belrose2023eliciting}; the naive prior is $\kappa = 0.5$.
    \item \textbf{(M5) Attention allocation}: $\alpha(\ell)$, the fraction of attention from the response-token position directed to manually annotated deep-constraint token spans, averaged across heads.
\end{itemize}
\subsection{Statistical Framework}\label{sec:stats}
All pairwise comparisons (deep-constraint vs.\ neutral queries, correct-run vs.\ incorrect-run, within-domain vs.\ cross-domain) use the Mann-Whitney $U$ test with Benjamini-Hochberg false discovery rate correction \citep{benjamini1995controlling} across the layers of each model. Effect sizes are rank-biserial $r = U / (n_1 \cdot n_2)$. Logit-lens commitment asymmetry uses a one-sample $t$-test against $\kappa = 0.5$ at the minimum-$\kappa$ layer per model.
\subsection{Causal Localization}\label{sec:causal-localization}
To test whether a single layer causally produces the layerwise dynamics measured in M1--M5, we run activation patching. The method is the following: for each query, at each layer $\ell$, we replace the residual-stream input to layer $\ell$ in the incorrect run with the corresponding state from the correct run and measure the normalized recovery of the correct-token logit:

\begin{equation}
    e(\ell) = (\text{logit}_{\text{patched}} - \text{logit}_{\text{i-baseline}}) / (\text{logit}_{\text{c-baseline}} - \text{logit}_{\text{i-baseline}})
\end{equation}

$e(\ell)=1$ indicates full recovery, $e(\ell)=0$ no causal contribution, and $e(\ell) < 0$ indicates interference (the layer's correct-run state pushes the incorrect-run logit away from recovery). We apply patching on LLaMA-2 13B, Mistral 7B and StableLM-2 1.6B. The alignment between the collected source states and the module receiving the patch is validated for models exposing the \texttt{model.model.layers} block container (LLaMA-2, Mistral); StableLM-2 exposes a different container (\texttt{transformer.h}) and, as reported in Section~\ref{sec:patching}, produces a profile inconsistent with a valid single-layer patch, so we treat it as an artefact and exclude it. Qwen2 1.5B is excluded upstream because its tokenizer fragments the response tokens into multiple sub-tokens for several queries, so the last-token-position extraction reads from an intermediate sub-token rather than from the response prediction. The implementation note on \texttt{forward\_pre\_hook} use under \texttt{transformers} $\geq 4.40$, the architecture-specific source-alignment issue, and the value-clipping protocol are in Appendix~\ref{app:patching}.
\section{Results}\label{sec:results}
\subsection{Isometric Rotational Divergence}\label{sec:rotational}
Table~\ref{tab:displacement} reports the displacement decomposition. Across the six models with measurable factual processing, mean displacement cosine similarity $\cos(d_\ell^+, d_\ell^-)$ drops from $1.00$ at the embedding layer to $0.65$--$0.80$ at intermediate depth, while mean norm ratios $\bar{\eta}(\ell)$ remain close to unity throughout (Table~\ref{tab:displacement}; quantified below). The model does not distinguish correct from incorrect by making one pathway louder; at the population level the discriminative signal is encoded in the angular component of the displacement.
\begin{table}[h]
\centering
\caption{\textbf{Norm ratio $\bar{\eta}(\ell)$ across models and layer bands.} Cells report mean~$\pm$~standard deviation of the per-query norm ratio $\eta_i(\ell) = \|d^{+}_{i,\ell}\| / \|d^{-}_{i,\ell}\|$, computed over $n=300$ queries per cell at the layer closest to each target fraction $\ell/L$. The mean stays close to unity at every cell across six architectures spanning $1$B to $13$B parameters, with the largest deviation at $5.5\%$ (Gemma 2 2B, final layer), supporting a population-level isometric divergence (Definition~\ref{def:isometric}). Qwen2 1.5B is reported separately: the bulk of its queries are excluded due to the tokenizer-extraction issue (see text), so the reduced SD reflects sample restriction, not lower spread.}
\label{tab:displacement}
\small
\setlength{\tabcolsep}{4pt}
\begin{tabular}{lcccc}
\toprule
\textbf{Model} & $\ell/L \approx 0.25$ & $\ell/L \approx 0.50$ & $\ell/L \approx 0.75$ & $\ell/L \approx 1.0$ \\
\midrule
Llama 3.2 1B          & $1.007 \pm 0.045$ & $1.006 \pm 0.062$ & $1.003 \pm 0.113$ & $1.027 \pm 0.111$ \\
StableLM 2 1.6B       & $1.009 \pm 0.051$ & $0.996 \pm 0.066$ & $0.990 \pm 0.099$ & $1.022 \pm 0.103$ \\
Gemma 2 2B            & $1.012 \pm 0.034$ & $0.996 \pm 0.070$ & $0.993 \pm 0.115$ & $1.055 \pm 0.218$ \\
Llama 3.2 3B          & $1.005 \pm 0.059$ & $1.010 \pm 0.068$ & $0.998 \pm 0.098$ & $1.019 \pm 0.077$ \\
Mistral 7B            & $1.004 \pm 0.073$ & $0.999 \pm 0.082$ & $1.000 \pm 0.097$ & $1.014 \pm 0.100$ \\
LLaMA-2 13B           & $0.999 \pm 0.060$ & $0.992 \pm 0.083$ & $0.995 \pm 0.108$ & $1.022 \pm 0.092$ \\
\midrule
Qwen2 1.5B (excluded) & $0.998 \pm 0.025$ & $0.998 \pm 0.025$ & $0.998 \pm 0.025$ & $0.998 \pm 0.025$ \\
\bottomrule
\end{tabular}
\end{table}
The mean across queries, $\bar{\eta}(\ell)$, stays close to unity at every layer band of every non-excluded model (Table~\ref{tab:displacement}): the largest mean deviation in any cell is $5.5\%$ (Gemma 2 2B at the final layer), and typical deviations are below $2\%$. Per-query spread is moderate at early and mid layers (SD typically $0.05$--$0.10$) and grows at the final layer (SD up to $0.22$ in Gemma 2 2B), but the mean remains near unity throughout: equal magnitudes are a property of the aggregate over queries---a population-level regularity---not of every individual query.
The consequence is geometric. At the population mean, where $|\bar{\eta}(\ell) - 1| < \delta$ (Definition~\ref{def:isometric}, Appendix~\ref{app:supplementary}), the squared distance between correct and incorrect representations is dominated by the angular term $2\,\bar{r}_\ell^{\,2}(1 - \cos\theta_\ell)$; the magnitude term contributes only to the extent that individual queries depart from $\eta_i = 1$. The discriminative signal is therefore carried principally by the angle between displacement vectors rather than by their lengths, though at the level of an individual query a magnitude contribution remains.
\subsection{Asymmetric Logit-Lens Commitment in the Incorrect Run}\label{sec:asymmetry}
\begin{figure}[h]
  \centering
  \includegraphics[width=\textwidth]{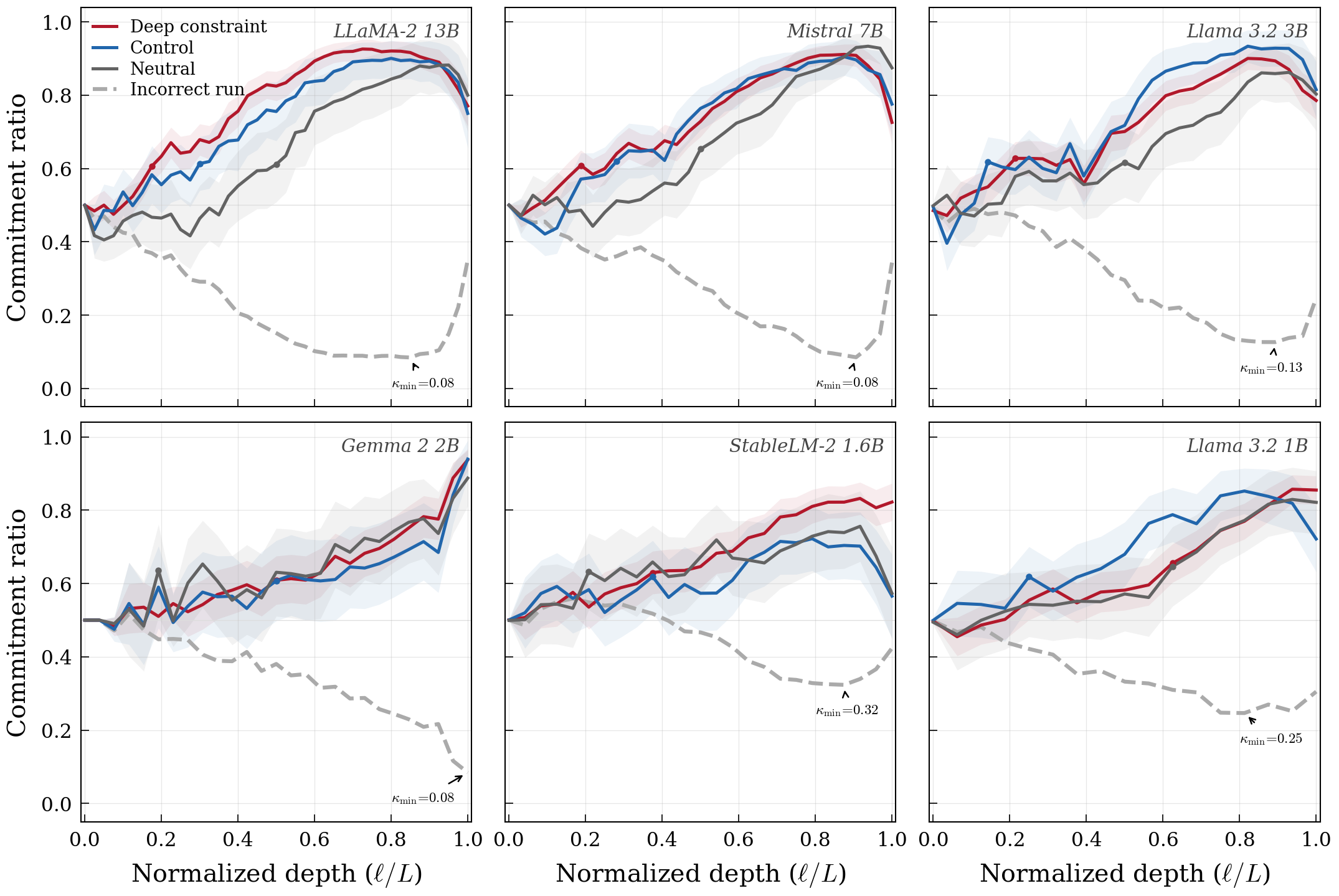}
  \caption{Logit-lens commitment dynamics across the six models with measurable factual processing ($2 \times 3$ grid, ordered large to small). Solid lines: $\kappa(\ell)$ during correct-run processing, stratified by query category (DC red, CTRL blue, NEU grey). Dashed grey line: mean $\kappa(\ell)$ during incorrect-run processing. Horizontal line at $\kappa = 0.5$ marks the naive prior of equal probability assignment. Annotated $\kappa_{\min}$ values indicate the minimum commitment reached during incorrect-run processing. Qwen2 1.5B (not shown) holds $\kappa \equiv 0.50$ throughout under our extraction protocol; see Section~\ref{sec:limitations}.}
  \label{fig:commitment}
\end{figure}

Figure~\ref{fig:commitment} shows the logit-lens commitment ratio $\kappa(\ell)$ across normalized depth for the six measurable models; Table~\ref{tab:commitment} reports per-model summary statistics with significance tests against the naive prior $\kappa = 0.5$. Solid colored lines show $\kappa$ during the correct run, stratified by query category (deep constraint red, control blue, neutral grey); the dashed grey line shows the mean value of $\kappa$ during the incorrect run.
In LLaMA-2 13B, Mistral 7B and Llama 3.2 3B, correct-run commitment rises from $0.50$ at the embedding layer to $0.76$--$0.79$ at the final layer (Table~\ref{tab:commitment}), with deep-constraint queries (red) leading control (blue) throughout; across all six measurable models the final-layer correct-run commitment spans $0.72$ (StableLM-2) to $0.93$ (Gemma 2). The incorrect-run dashed line falls to $\kappa_{\min} = 0.08$ in LLaMA-2 13B and Mistral 7B and $0.13$ in Llama 3.2 3B. In Gemma 2 2B, the incorrect-run $\kappa_{\min} = 0.08$ matches LLaMA-2 13B despite an order-of-magnitude smaller parameter count. StableLM-2 1.6B reaches $\kappa_{\min} = 0.32$; Llama 3.2 1B reaches $\kappa_{\min} = 0.25$.
\begin{table}[h]
\centering
\caption{Logit-lens commitment dynamics across the seven models. $\bar{\kappa}^+_{\text{final}}$: mean commitment at the final layer during correct-run processing. $\bar{\kappa}^-_{\min}$: minimum mean commitment during incorrect-run processing. Suppression depth $\sigma = 0.5 - \bar{\kappa}^-_{\min}$. One-sample $t$-test against $\kappa = 0.5$ at the minimum-$\kappa$ layer; $n = 300$.}
\label{tab:commitment}
\small
\begin{tabular}{@{}lcccccc@{}}
\toprule
Model & $\bar{\kappa}^+_{\text{final}}$ & $\bar{\kappa}^-_{\min}$ & $\ell^*/L$ & $\sigma$ & $t$ & $p$ \\
\midrule
LLaMA-2 13B       & 0.77 & 0.08 & 0.85 & 0.42 & $-36.5$ & $< 10^{-100}$ \\
Mistral 7B        & 0.76 & 0.08 & 0.91 & 0.42 & $-39.1$ & $< 10^{-100}$ \\
Gemma 2 2B        & 0.93 & 0.08 & 1.00 & 0.42 & $-30.1$ & $< 10^{-91}$ \\
Llama 3.2 3B      & 0.79 & 0.13 & 0.93 & 0.37 & $-25.0$ & $< 10^{-74}$ \\
StableLM-2 1.6B   & 0.72 & 0.32 & 0.88 & 0.18 & $-7.9$  & $< 10^{-13}$ \\
Llama 3.2 1B      & 0.82 & 0.25 & 0.87 & 0.25 & $-12.8$ & $< 10^{-29}$ \\
Qwen2 1.5B        & 0.50 & 0.50 & --   & 0.00 & $-0.01$ & $0.99$ \\
\bottomrule
\end{tabular}
\end{table}
The observed $\kappa_{\min} = 0.08$ corresponds to the model assigning the incorrect token approximately $11.5$ times the probability of the correct token, with $\kappa$ in the incorrect run far below the naive prior of $0.5$. The asymmetry is consistent across the four measurable architecture families (Llama, Mistral, Gemma, StableLM) and replicates within architecture families with scale.
\subsection{Linear Probe Accuracy}\label{sec:probes}
\begin{figure}[h]
  \centering
  \includegraphics[width=\textwidth]{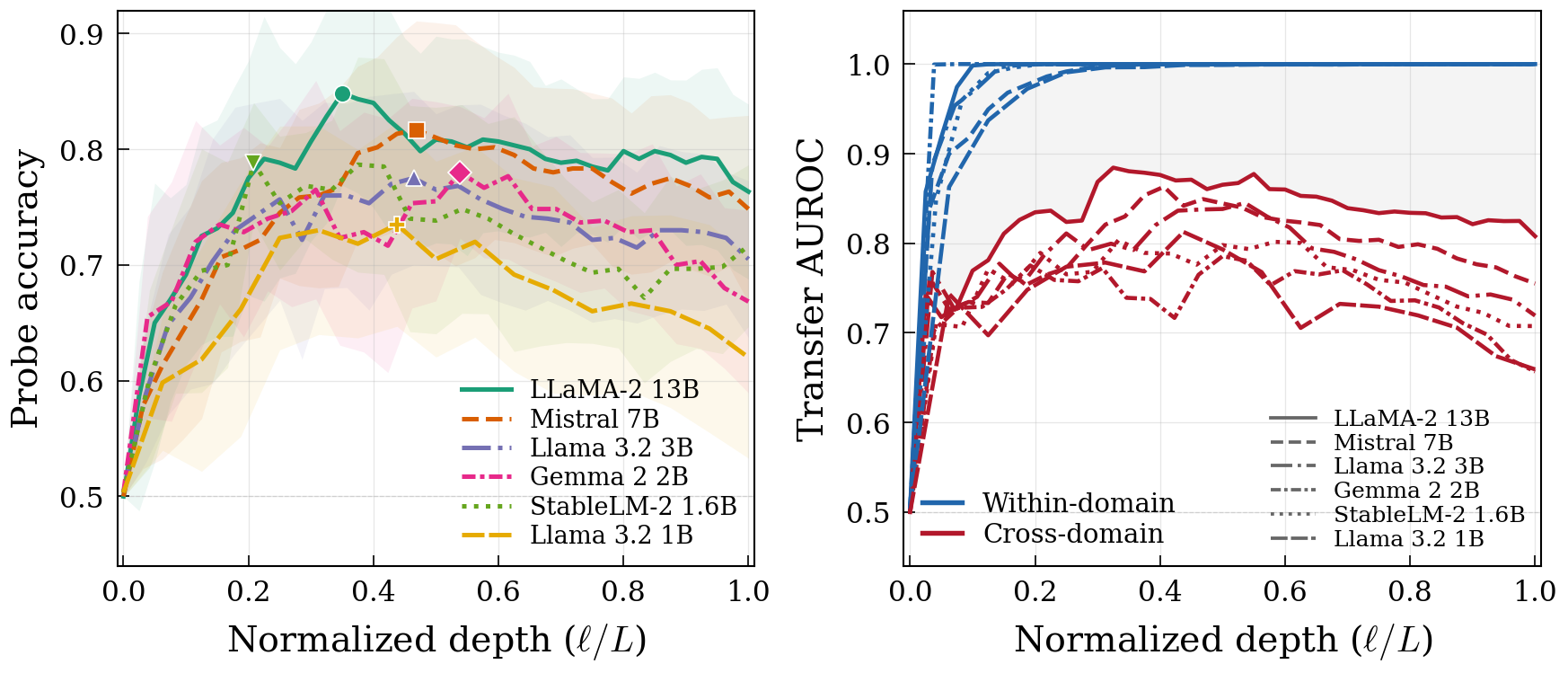}
  \caption{Probe accuracy and cross-domain transfer across the six measurable models. \textbf{Left:} Linear probe accuracy (5-fold CV, shaded $\pm 1$ SD) across normalised depth. All models peak at intermediate depth; within the Llama family the peak shifts deeper and rises with parameter count. \textbf{Right:} Cross-domain transfer AUROC. Blue lines: within-domain (near-perfect for all models from early mid-layers onward). Red lines: cross-domain. The financial--medical corridor (upper red lines) transfers consistently better than transport-domain pairs (lower red lines), a structural asymmetry replicated across all six models.}
  \label{fig:transfer}
\end{figure}

We observe in Figure~\ref{fig:transfer} that all six measurable models rise from chance at the embedding layer to a peak at intermediate depth ($\ell/L \in [0.21, 0.52]$) and then decline toward the final layer; within the Llama family peak accuracy scales with parameter count ($0.74$ at 1B, $0.78$ at 3B, $0.85$ at 13B), and Gemma 2 2B peaks at $0.78$ matching the 3B Llama. The post-peak decline of $\Delta = 0.07$--$0.12$ is consistent across all six models. Full per-model probe accuracies are in Table~\ref{tab:probes} (Appendix~\ref{app:supplementary}). The qualitative shape is consistent with prior layerwise probe work \citep{alain2017understanding, belrose2023eliciting}.
\subsection{Attention Reallocates to Deep-Constraint Tokens in Correct Runs}\label{sec:attention}
The attention allocation ratio $\alpha(\ell)$ (M5) to manually annotated deep-constraint tokens, averaged across heads at the response-token position, is systematically higher in the correct run than in the incorrect run (Table~\ref{tab:attention}, Appendix~\ref{app:supplementary}). The per-model mean across queries and layers reaches $\bar{\alpha}^{+} = 0.60$--$0.65$ in correct runs versus $\bar{\alpha}^{-} = 0.42$--$0.46$ in incorrect runs. The $\approx 0.15$--$0.20$ gap is significant at 15--39 of the available layers per model under FDR-corrected Mann-Whitney $U$ testing, with rank-biserial correlations $r_{\max} = 0.35$--$0.43$ across the six measurable models. Llama 3.2 1B reaches $r_{\max} = 0.42$ on attention while showing only moderate $\kappa$ asymmetry. We report this correlation as an observation, not as evidence that attention reallocation causes the commitment asymmetry.
\subsection{Cross-Domain Probe Transfer and Structural Asymmetry}\label{sec:transfer}
We train linear probes on hidden states from one domain and test on each of the other two. The right panel of Figure~\ref{fig:transfer} shows transfer AUROC across normalised depth.
Within-domain AUROC is near-perfect for all six models from early mid-layers onward. Cross-domain transfer is consistently weaker and structurally asymmetric: financial--medical AUROC reaches $0.80$--$0.96$ at the final layer, while transport-domain transfer to or from either other domain yields $0.52$--$0.74$. The transport-domain asymmetry replicates across all six models, which makes a model-specific artefact unlikely; we read it as reflecting the shared regulatory-numeric structure of the financial and medical stems relative to the transport stems, and note that we do not have an independent measure of pretraining co-occurrence to confirm that reading.
\subsection{Single-Layer Activation Patching}\label{sec:patching}
On LLaMA-2 13B and Mistral 7B---the two models where the patch is cleanly interpretable---single-layer activation patching produces no layer band where patching from the correct run recovers the correct token in the incorrect run. LLaMA-2 13B shows mean per-layer effect $\approx 0$ across depth bands and category types. Mistral 7B shows a small, high-variance category-conditional effect on deep-constraint queries (mean $0.32 \pm 0.96$) but not on control or neutral, and the effect does not persist across the layer band; given the standard deviation it does not support a localized site.
StableLM-2 1.6B instead recovers uniformly at every layer---including the first---with mean effect at or above full recovery ($\approx 1.0$, and $1.38$ on control queries, i.e.\ beyond the recovery ceiling of $1$). A valid single-layer patch cannot recover the answer from the input to the first layer, and normalized recovery cannot legitimately exceed $1$; this profile is diagnostic of an invalid patch on the StableLM code path (Appendix~\ref{app:patching}) rather than of localization. We therefore exclude StableLM-2 from the causal analysis pending a corrected re-run, and report a \emph{scoped null}: within the two cleanly-patched architectures, single-layer patching does not localize the late-layer asymmetry. Per-layer effect tables are in Appendix~\ref{app:patching}.
\section{Discussion}\label{sec:discussion}
\subsection{Scope of the geometric claim}
In the incorrect run, $\kappa_{\min} = 0.08$ means the model assigns the appended incorrect token roughly $11.5\times$ the probability of the correct one at the response position. The matched-pair design controls for syntactic position and domain-meaningful framing of $t^{+}$ and $t^{-}$; under that design, a model treating both continuations as equally valid completions would produce $\kappa \approx 0.5$ in both runs. The observed asymmetry is therefore difficult to explain by neutral context-following alone. Two readings remain compatible with the incorrect-run trajectory: the model rejects the wrong continuation, or it conforms to the token it is forced to carry. Our claim rests on the divergence between the matched runs rather than on the incorrect run alone; separating rejection from conformity is exactly what a random-token baseline---a third forced run with a frequency-matched but semantically unrelated token---would do, and it is the priority next experiment (Section~\ref{sec:limitations}). The asymmetry calls for a causal explanation, but its mechanism remains open: the scoped null on single-layer patching (Section~\ref{sec:patching}) bounds what causal account our data support on the two cleanly-patched architectures, and we cannot decide between a mechanism distributed across many small contributions and a localized mechanism that single-layer patching fails to detect. Discriminating these accounts further requires span-based patching across consecutive layers and direction-based interventions on identified residual-stream subspaces. We do not claim that our measurements characterize hallucination as it arises in free generation; the protocol is a controlled probe under forced continuations, and whether identical dynamics arise under unconstrained generation is the central empirical question we leave open.
\subsection{Implications for measurement and detection}
Four implications follow within the forced-completion regime and require verification under free-generation conditions before being extended. (i) Because divergence is isometric at tolerance $\delta \leq 0.06$ at the population mean (Definition~\ref{def:isometric}, Table~\ref{tab:displacement}), methods that compare only displacement magnitudes will miss the primary signal; direction-based metrics are the relevant tool. (ii) Because $\kappa$ asymmetry forms through mid-to-late depth rather than at the output, layerwise probes access information that output-layer detection cannot. (iii) Because cross-domain probe transfer is structurally asymmetric, per-domain calibration is prudent; the financial--medical corridor transfers far better than transport pairs, a regularity consistent with shared regulatory-numeric stem structure and with regional and topological accounts \citep{steenrod1951, bronstein2021geometric}, but not adjudicated by our data. (iv) Because within the scales tested architecture rather than parameter count alone appears to set the suppression-depth ceiling---Gemma 2 2B reaches $\sigma = 0.42$ at 2B parameters, matching LLaMA-2 13B at 13B and exceeding the equivalent-scale Llama 3.2 3B ($\sigma = 0.37$)---within-family parameter sweeps are needed to separate the architectural contribution from scale. Candidate architectural choices include Gemma 2's alternating local/global attention and logit softcapping \citep{gemmateam2024}; we flag these as hypotheses, not conclusions, since a four-family sample cannot isolate a single mechanism.
\section{Limitations}\label{sec:limitations}
\paragraph{Scope of inference.}
The protocol observes hidden-state geometry under matched forced continuations, not under unconstrained autoregressive generation; we do not claim identity of dynamics across regimes. Free-generation experiments on the same query stems with model-produced hallucinations are the priority extension.
\paragraph{Causal protocols.}
Two controls are missing. (i) A random-token baseline (a third forced run with a tokenizer-frequency-matched but semantically unrelated continuation) would discriminate sensitivity to factual mismatch from general accommodation of any forced continuation---the rejection-vs-conformity ambiguity noted in Section~\ref{sec:discussion}. (ii) A query-only logit-lens trajectory would measure intermediate preference for the correct token before any continuation is forced. Together with span-based patching, direction-based interventions on residual-stream subspaces, resample ablations, and a corrected patching hook for the StableLM code path, they would discriminate distributed from localized-but-undetected mechanistic accounts.
\paragraph{Patching scope.}
Single-layer patching was run on three of the seven models and is cleanly interpretable on two (LLaMA-2 13B, Mistral 7B); the StableLM-2 result is excluded as an artefact of its block-container code path (Appendix~\ref{app:patching}). The scoped null therefore rests on two architectures of the same broad family and should not be read as a claim about all seven models or all architectures.
\paragraph{Dataset and model scope.}
The 300 queries are human-authored under a fixed protocol by a single annotator; inter-annotator reliability on a 50-query partial re-categorization is planned. The forced-completion protocol assumes a single BPE response token; for Qwen2 1.5B several financial/medical responses fragment into sub-tokens, so the last-token extraction reads from a non-response position---this explains the flat $\kappa$ and means Qwen2 1.5B's characterization as a scale floor is not safely supported until per-tokenizer extraction is corrected. Five architecture families across seven models do not separate scale from architectural effects cleanly; within-family parameter sweeps are needed, and encoder--decoder and retrieval-augmented architectures may differ.
\section{Conclusions}\label{sec:conclusions}
We introduced forced-completion probing as a controlled instrument for measuring layerwise hidden-state geometry of factual queries, applied it to seven decoder-only transformers across five architecture families, and documented a near-isometric rotational divergence pattern together with an asymmetric late-layer commitment, alongside a scoped null result for single-layer causal localization on the two architectures where patching is cleanly interpretable. The geometric structure is observable and robust across six measurable models; the mechanism that produces it remains open, as does whether identical dynamics arise under unconstrained autoregressive generation, and whether the incorrect-run trajectory reflects rejection or conformity. We release the protocol, the 300-query annotated dataset, and the measurement battery to enable the random-token, query-only, corrected-patching and span/direction-based intervention experiments that would close these questions.
\section*{Reproducibility}
The dataset, measurement code, and complete extraction protocol (including the per-tokenizer response-position localization and the architecture-specific patching-hook alignment discussed in Section~\ref{sec:limitations} and Appendix~\ref{app:patching}) are available under request to author. All experiments were run on a single NVIDIA A100 40\,GB. Per-query, per-layer, per-model raw measurements are released as CSV files; aggregation scripts that produce the tables and figures in this paper are released as Jupyter notebooks.
\bibliographystyle{plainnat}
\bibliography{references}
\appendix
\section{Visual Summary}\label{app:figure}
Figure~\ref{fig:central_pattern} summarizes the joint observational pattern documented in Sections~\ref{sec:rotational} and \ref{sec:asymmetry}. We show this figure to highlight two simultaneous regularities. First, in Panel~A, the dashed concentric arcs make explicit that the four displacement vectors at successive normalized depths terminate at approximately equal radii in the correct and incorrect runs---the radial differences between the two pathways are small relative to the angular separation at every depth. The discriminative signal is encoded in the angle $\theta(\ell)$, not in the magnitudes. Second, in Panel~B, the dashed horizontal at $\kappa=0.5$ is the naive prior an observer would expect if the residual stream merely accommodated the appended token without preference. The $\kappa_{\min}=0.08$ achieved in the incorrect run corresponds to the model placing approximately $11.5\times$ more probability mass on the incorrect token than on the correct one---an asymmetry whose mechanistic explanation remains open under our scoped single-layer patching null (Section~\ref{sec:patching} and Appendix~\ref{app:patching}).
\begin{figure}[h]
\centering
\begin{tikzpicture}[
  >={Latex[length=2.2mm, width=1.6mm]},
  scale=0.9,
  font=\small\sffamily,
]
\begin{scope}[shift={(0,0)}]
  \node[anchor=south west, font=\sffamily\small\bfseries, querygray]
    at (-3.5, 4.5) {(A)\;\, Hidden-state displacement geometry};
  \foreach \r/\op in {1.1/15, 2.0/20, 2.9/25, 3.8/30} {
    \draw[gray!\op, very thin, dashed] (0,0) circle (\r);
  }
  \node[font=\tiny, querygray] at (1.1, -0.25) {$\ell/L = 0.1$};
  \node[font=\tiny, querygray] at (2.0, -0.25) {0.35};
  \node[font=\tiny, querygray] at (2.9, -0.25) {0.65};
  \node[font=\tiny, querygray] at (3.8, -0.25) {1.0};
  \filldraw[querygray] (0,0) circle (2.2pt);
  \node[anchor=north east, querygray, font=\small] at (0.05, -0.05) {$h_\ell^{\,q}$};
  \draw[->, correctgreen, line width=0.55mm] (0,0) -- ({1.1*cos(88)}, {1.1*sin(88)});
  \draw[->, correctgreen, line width=0.65mm] (0,0) -- ({2.0*cos(80)}, {2.0*sin(80)});
  \draw[->, correctgreen, line width=0.75mm] (0,0) -- ({2.9*cos(72)}, {2.9*sin(72)});
  \draw[->, correctgreen, line width=0.85mm] (0,0) -- ({3.8*cos(64)}, {3.8*sin(64)});
  \draw[->, incorrectred, line width=0.55mm] (0,0) -- ({1.1*cos(92)}, {1.1*sin(92)});
  \draw[->, incorrectred, line width=0.65mm] (0,0) -- ({2.0*cos(100)}, {2.0*sin(100)});
  \draw[->, incorrectred, line width=0.75mm] (0,0) -- ({2.9*cos(108)}, {2.9*sin(108)});
  \draw[->, incorrectred, line width=0.85mm] (0,0) -- ({3.8*cos(116)}, {3.8*sin(116)});
  \draw[anglecolor, line width=0.35mm, ->, >={Latex[length=1.6mm]}]
    ({3.55*cos(64)}, {3.55*sin(64)}) arc[start angle=64, end angle=116, radius=3.55];
  \node[anglecolor, font=\small]
    at ({3.55*cos(90)+0.20}, {3.55*sin(90)+0.35}) {$\theta(\ell)$};
  \node[correctgreen, anchor=west, font=\small\bfseries]
    at ({3.8*cos(64)+0.15}, {3.8*sin(64)-0.05}) {$d^{+}_\ell$ (correct)};
  \node[incorrectred, anchor=east, font=\small\bfseries]
    at ({3.8*cos(116)-0.15}, {3.8*sin(116)-0.05}) {$d^{-}_\ell$ (incorrect)};
  \node[anchor=north, font=\footnotesize, querygray, align=center, text width=8.2cm]
    at (0, -0.8) {Equal-magnitude displacement vectors ($\|d^{+}\|/\|d^{-}\|\!\approx\!1$)\\
                  rotate apart with depth (cosine drops, magnitudes stay)};
\end{scope}
\begin{scope}[shift={(-4, -11.0)}]
  \node[anchor=south west, font=\sffamily\small\bfseries, querygray]
    at (-0.7, 5.7) {(B)\;\, Logit-lens commitment $\kappa(\ell)$};
  \draw[->, querygray, thin] (0,0) -- (6.7, 0) node[right, font=\small, querygray] {$\ell/L$};
  \draw[->, querygray, thin] (0,0) -- (0, 5.4) node[above, font=\small, querygray] {$\kappa$};
  \node[anchor=east, font=\tiny, querygray] at (0, 0.4)  {0.08};
  \node[anchor=east, font=\tiny, querygray] at (0, 2.5)  {0.5};
  \node[anchor=east, font=\tiny, querygray] at (0, 3.85) {0.77};
  \node[anchor=east, font=\tiny, querygray] at (0, 5.0)  {1.0};
  \foreach \x/\lbl in {0/0, 1.625/0.25, 3.25/0.5, 4.875/0.75, 6.5/1.0} {
    \draw[querygray] (\x, -0.05) -- (\x, 0.05);
    \node[font=\tiny, querygray, below] at (\x, -0.05) {\lbl};
  }
  \draw[gray!60, dashed, very thin] (0, 2.5) -- (6.5, 2.5);
  \node[querygray, font=\tiny, anchor=south west] at (0.15, 2.55) {naive prior $\kappa=0.5$};
  \draw[correctgreen, line width=0.6mm, smooth]
    plot coordinates {
      (0, 2.5) (0.5, 2.55) (1, 2.65) (1.6, 2.85)
      (2.2, 3.10) (2.8, 3.35) (3.4, 3.55) (4.0, 3.70)
      (4.6, 3.78) (5.2, 3.82) (5.8, 3.84) (6.5, 3.85)
    };
  \draw[incorrectred, line width=0.6mm, smooth]
    plot coordinates {
      (0, 2.5) (0.5, 2.45) (1, 2.35) (1.6, 2.15)
      (2.2, 1.85) (2.8, 1.45) (3.4, 1.05) (4.0, 0.70)
      (4.6, 0.50) (5.2, 0.42) (5.8, 0.40) (6.5, 0.44)
    };
  \filldraw[correctgreen] (6.5, 3.85) circle (1.2pt);
  \node[correctgreen, anchor=west, font=\tiny] at (6.5, 3.85) {\,0.77};
  \filldraw[incorrectred] (5.8, 0.40) circle (1.2pt);
  \node[incorrectred, anchor=south, font=\tiny] at (5.4, 0.50) {$\kappa_{\min}=0.08$};
  \node[correctgreen, font=\small\bfseries, anchor=west] at (3.7, 4.30) {correct run};
  \node[incorrectred, font=\small\bfseries, anchor=west] at (3.7, 1.30) {incorrect run};
  \begin{scope}[on background layer]
    \fill[anglecolor!10]
      plot[smooth] coordinates {(2.8, 3.35) (3.4, 3.55) (4.0, 3.70) (4.6, 3.78) (5.2, 3.82) (5.8, 3.84) (6.5, 3.85)}
      -- (6.5, 0.44)
      -- plot[smooth] coordinates {(6.5, 0.44) (5.8, 0.40) (5.2, 0.42) (4.6, 0.50) (4.0, 0.70) (3.4, 1.05) (2.8, 1.45)}
      -- cycle;
  \end{scope}
  \draw[decorate, decoration={brace, amplitude=4pt, mirror}, anglecolor, thin]
    (6.7, 0.44) -- (6.7, 3.85);
  \node[anglecolor, font=\footnotesize, anchor=west, align=left]
    at (6.95, 2.15) {$11.5\times$\\probability\\asymmetry};
  \node[anchor=north, font=\footnotesize, querygray, align=center, text width=7.5cm]
    at (3.25, -0.55) {Late layers resolve into asymmetric preference:\\
                       $\kappa$ in the incorrect run falls far below $0.5$};
\end{scope}
\end{tikzpicture}
\caption{Visual summary of the central observational pattern. \textbf{(A)} In displacement space, correct- and incorrect-run vectors from the query-only representation $h_\ell^{\,q}$ rotate apart with depth while maintaining approximately equal magnitude ($\bar{\eta}(\ell) \approx 1$; see Table~\ref{tab:displacement}). The angular separation $\theta(\ell)$ grows through mid-depth before partial reconvergence at the final layer. \textbf{(B)} Late layers resolve the asymmetric outcome: the logit-lens commitment ratio $\kappa(\ell)$ rises above the naive prior of $0.5$ in the correct run and falls far below it in the incorrect run, reaching $\kappa_{\min} = 0.08$ in the largest models (Table~\ref{tab:commitment}). The asymmetry corresponds to the model assigning the incorrect token approximately $11.5\times$ the probability of the correct token at the minimum-$\kappa$ layer.}
\label{fig:central_pattern}
\end{figure}
\section{Supplementary Tables and Definitions}\label{app:supplementary}
This appendix collects per-model numerical tables omitted from the main text for space, and the formal definition of \emph{isometric divergence} referenced from Section~\ref{sec:rotational}.
\begin{definition}[Isometric divergence]\label{def:isometric}
A pair of forced runs exhibits \emph{isometric divergence} at layer $\ell$ \emph{at tolerance} $\delta$ if $|\bar{\eta}(\ell) - 1| < \delta$. Under this condition, the squared distance between the mean correct and incorrect representations satisfies
\begin{equation}
\|h_\ell^+ - h_\ell^-\|^2 \;=\; 2\,\bar{r}_\ell^{\,2}\,(1 - \cos\theta_\ell) + O(\delta),
\end{equation}
where $\theta_\ell$ is the angle between displacement vectors and $\bar{r}_\ell$ is the mean displacement magnitude across the two runs. At the population mean the discriminative signal resides predominantly in $\theta_\ell$ rather than in the magnitudes; the magnitude contribution vanishes as $\delta \to 0$. The statement is a population-mean identity: for an individual query with $\eta_i \neq 1$, a magnitude contribution remains.
In our data (Table~\ref{tab:displacement}), isometric divergence holds at tolerance $\delta \leq 0.06$ across all cells.
\end{definition}
\begin{table}[h]
\centering
\caption{Linear probe accuracy (5-fold cross-validation). All six measurable models peak at intermediate depth with a consistent post-peak decline.}
\label{tab:probes}
\small
\begin{tabular}{@{}lcccc@{}}
\toprule
Model & Peak accuracy & Peak $\ell/L$ & Final accuracy & $\Delta$ \\
\midrule
LLaMA-2 13B     & $0.85 \pm 0.08$ & 0.35 & 0.76 & 0.09 \\
Mistral 7B      & $0.82 \pm 0.09$ & 0.47 & 0.75 & 0.07 \\
Llama 3.2 3B    & $0.78 \pm 0.08$ & 0.45 & 0.71 & 0.07 \\
Gemma 2 2B      & $0.78 \pm 0.06$ & 0.52 & 0.67 & 0.11 \\
StableLM-2 1.6B & $0.79 \pm 0.05$ & 0.21 & 0.72 & 0.07 \\
Llama 3.2 1B    & $0.74 \pm 0.09$ & 0.41 & 0.62 & 0.12 \\
Qwen2 1.5B      & $0.50$          & --   & 0.50 & 0.00 \\
\bottomrule
\end{tabular}
\end{table}
\begin{table}[h]
\centering
\caption{Attention allocation to deep-constraint tokens across the seven models. Correct runs systematically allocate more attention to factual anchor tokens. FDR-corrected Mann-Whitney $U$ test across layers; $r$ is rank-biserial correlation.}
\label{tab:attention}
\small
\begin{tabular}{@{}lccccc@{}}
\toprule
Model & $\bar{\alpha}^+$ & $\bar{\alpha}^-$ & Sig.\ layers & $r_{\max}$ & Best $p_{\text{FDR}}$ \\
\midrule
LLaMA-2 13B       & 0.62 & 0.46 & 39/41 & 0.35 & $< 10^{-5}$ \\
Mistral 7B        & 0.60 & 0.42 & 31/33 & 0.35 & $< 10^{-5}$ \\
Llama 3.2 3B      & 0.64 & 0.45 & 28/29 & 0.41 & $< 10^{-6}$ \\
Gemma 2 2B        & 0.63 & 0.42 & 21/27 & 0.43 & $< 10^{-6}$ \\
StableLM-2 1.6B   & 0.61 & 0.42 & 18/25 & 0.40 & $< 10^{-6}$ \\
Llama 3.2 1B      & 0.65 & 0.42 & 15/17 & 0.42 & $< 10^{-5}$ \\
Qwen2 1.5B        & 0.48 & 0.34 & 0/29  & 0.07 & 0.20 \\
\bottomrule
\end{tabular}
\end{table}
\section{Dataset Construction}\label{app:dataset}
This appendix expands the summary in Section~\ref{sec:dataset}. The dataset is 300 human-authored queries, stratified across three professional domains ($n=100$ each) and three category types under a fixed authoring protocol.
\paragraph{Deep constraint (DC, $n=182$).} The correct answer requires integrating a domain-specific factual rule with the query context. The query stem contains both a \emph{deep-constraint cue} (factual content that determines the correct answer under domain knowledge) and a \emph{surface cue} (a syntactic or numerical pattern that, taken alone, would favor the incorrect answer). The per-domain DC categories are:
\begin{itemize}
\item \textbf{Financial compliance:} \emph{RA} (risk-aware suitability: retirement-age portfolio rules, conservative-investor risk profiles) and \emph{RC} (regulatory conflict: leverage prohibitions for conservative investors, suitability under FINRA/MiFID-style frameworks).
\item \textbf{Medical protocols:} \emph{DI} (drug interactions: e.g., warfarin-aspirin, SSRI-NSAID) and \emph{CI} (contraindications: e.g., methotrexate-NSAID, beta-blocker--asthma).
\item \textbf{Transport regulation:} \emph{FD} (functional dependency: the object must be at the destination for the task, e.g., taking a car to a car wash) and \emph{PT} (physical transport: object too heavy or bulky to carry on foot regardless of distance).
\end{itemize}
\paragraph{Control (CTRL, $n=66$).} The correct answer is determinate and the surface and deep cues agree. The protocol matches the deep constraint categories in length and surface complexity but removes the surface-vs-deep tension. Per-domain categories are \emph{FC} (financial), \emph{MC} (medical), and \emph{TD} (transport---distance is the legitimate basis for the decision).
\paragraph{Neutral (NEU, $n=52$).} The correct/incorrect distinction does not depend on domain-specific factual reasoning, serving as a within-protocol null control following \citet{hewitt2019designing}. Per-domain categories are \emph{FN}, \emph{MN}, \emph{ND}.
\paragraph{Authoring constraints.} Each query is human-authored under three constraints: (i) exactly one single-token correct continuation $t^+$ and exactly one single-token incorrect continuation $t^-$, both domain-meaningful and not arbitrary distractors; (ii) for the studied tokenizers, both $t^+$ and $t^-$ map to single BPE tokens (per-model tokenizer constraints in Section~\ref{sec:limitations}); (iii) deep-keyword and surface-keyword annotations are character-offset-aligned to the query string for attention analysis.
\paragraph{Example query (RA, financial domain).} \texttt{Query:} ``Client is 68 and retired with \$500K savings. High-yield bonds at 8\% or treasury bonds at 4\%?'' \texttt{Correct:} ``Treasury bonds. High-yield bonds carry significant default risk inappropriate for a retiree's portfolio.'' \texttt{Incorrect:} ``High-yield bonds. Double the return means faster growth of retirement savings.'' \texttt{Deep keywords:} retired, default, risk, inappropriate, retiree. \texttt{Surface keywords:} 8\%, return, faster, growth.
\paragraph{Example query (DI, medical domain).} \texttt{Query:} ``Patient has a mild headache and is on warfarin. Should they take aspirin or acetaminophen?'' \texttt{Correct:} ``Acetaminophen. Aspirin interacts with warfarin and significantly increases bleeding risk.'' \texttt{Incorrect:} ``Aspirin. It is the standard and most effective headache treatment.'' \texttt{Deep keywords:} warfarin, interact, bleeding, aspirin. \texttt{Surface keywords:} mild, headache, standard.
\paragraph{Example query (FD, transport domain).} \texttt{Query:} ``I want to wash my car and the car wash is only 100m away. Should I drive or walk?'' \texttt{Correct:} ``Drive. You need the car at the car wash.'' \texttt{Incorrect:} ``Walk. It is only 100m away, no reason to drive such a short distance.'' \texttt{Deep keywords:} wash, car. \texttt{Surface keywords:} 100m, away.
The complete dataset, including per-query annotations of deep-keyword and surface-keyword character spans, is released alongside the paper.
\section{Activation Patching: Full Results and Protocol Note}\label{app:patching}
This appendix expands Section~\ref{sec:patching}. Table~\ref{tab:patching} reports the mean per-layer patching effect across queries, normalized-depth bands, and category types. The StableLM-2 row is retained for transparency but flagged as an artefact and excluded from interpretation (see below).
\begin{table}[h]
\centering
\caption{Single-layer activation patching results. Effect $e(\ell)$ is the normalized recovery of the correct-token logit when the residual stream at layer $\ell$ is patched from the correct into the incorrect run; $e = 1$ indicates full recovery, $e = 0$ no causal contribution, $e < 0$ that the layer's correct-run state interferes with the incorrect run. Values are mean $\pm$ standard deviation across queries within each bin. StableLM-2 1.6B is flagged as an artefact of its block-container code path (see protocol note) and excluded from interpretation.}
\label{tab:patching}
\footnotesize
\setlength{\tabcolsep}{3pt}
\begin{tabular}{@{}lcccccc@{}}
\toprule
Model & \multicolumn{3}{c}{Mean effect by depth band} & \multicolumn{3}{c}{Mean effect by category}\\
\cmidrule(lr){2-4}\cmidrule(lr){5-7}
& Early & Mid & Late & DC & CTRL & NEU \\
\midrule
LLaMA-2 13B                    & $\phantom{-}0.01 \pm 0.98$ & $-0.01 \pm 0.97$ & $-0.05 \pm 0.95$ & $\phantom{-}0.02 \pm 0.94$ & $-0.03 \pm 1.07$ & $-0.12 \pm 0.91$ \\
Mistral 7B                     & $\phantom{-}0.23 \pm 0.98$ & $\phantom{-}0.21 \pm 0.97$ & $\phantom{-}0.16 \pm 0.96$ & $\phantom{-}0.32 \pm 0.96$ & $\phantom{-}0.07 \pm 1.00$ & $-0.06 \pm 0.90$ \\
StableLM-2 1.6B \emph{(art.)}  & $\phantom{-}1.04 \pm 0.85$ & $\phantom{-}1.13 \pm 0.80$ & $\phantom{-}1.01 \pm 0.89$ & $\phantom{-}0.95 \pm 0.82$ & $\phantom{-}1.38 \pm 0.74$ & $\phantom{-}1.14 \pm 0.91$ \\
\bottomrule
\end{tabular}
\end{table}
\paragraph{Three observations.} First, in LLaMA-2 13B, mean per-layer effect is approximately zero across all depth bands and all category types. No layer or layer band emerges as a localized causal site for correct-token recovery in the incorrect run. Second, in Mistral 7B there is a small category-conditional effect: mean effect on deep-constraint queries is $0.32$ compared to $-0.06$ on neutral. The effect is in the expected direction but small in magnitude, high in variance ($\pm 0.96$), and inconsistent across the layer band, so it does not support a localized site. Third, in StableLM-2 1.6B mean patching effect is at or above full recovery ($\approx 1.0$, and $1.38$ on control) uniformly across layers---including the earliest---and across category types. Because a single-layer patch at the first layer cannot supply the downstream computation that produces the answer, and because normalized recovery cannot legitimately exceed $1$, this profile is not interpretable as localization; it is diagnostic of an invalid patch on the StableLM code path. We exclude StableLM-2 from the causal conclusion.
\paragraph{Interpretive bound.} We are deliberately conservative about the consequences of these results. The scoped null (on LLaMA-2 13B and Mistral 7B) does not establish that the late-layer asymmetry is distributed across many components rather than localized in a way our protocol fails to detect. Span-based patching (patching across multiple consecutive layers), direction-based interventions on identified residual-stream subspaces, and resample ablations would all be needed to discriminate between distributed-processing and localized-but-undetected accounts.
\paragraph{Implementation note on hook recursion.} We document the following implementation detail because earlier protocol versions failed silently. Under \texttt{transformers} $\geq 4.40$, setting \texttt{output\_hidden\_states=True} installs internal forward hooks on every layer to collect hidden states. If our patching protocol additionally installs \texttt{forward\_pre\_hook}s on the same layers, the result is mutual recursion between our hook and the internal collection hook, manifesting either as infinite-loop failure (caught by the Python recursion limit) or, more dangerously, as silent NaN output. The fix used in our protocol is to set \texttt{output\_hidden\_states=False} at the model call site and pass hidden-state collection requests transiently per forward pass. We install our pre-hook on one layer at a time with a hook handle that is always removed in a \texttt{finally} block, regardless of forward-pass exceptions. The released code documents this in inline comments.
\paragraph{Architecture-specific source alignment.} A second implementation detail explains the StableLM-2 artefact. Our patch replaces the residual-stream input to layer $\ell$ in the incorrect run with the source state collected at the corresponding index from the correct run. The index alignment between the collected hidden-state sequence and the module receiving the pre-hook was validated for models exposing the \texttt{model.model.layers} block container (LLaMA-2, Mistral, and the Llama 3.2 models). StableLM-2 exposes a different container (\texttt{transformer.h}); under that path the source index does not align to the same residual position, so the injected state carries near-final information at every layer, which produces the observed uniform, above-ceiling recovery. The corrected implementation we release captures the source and applies the patch through the identical \texttt{forward\_pre\_hook} mechanism, so alignment holds by construction across architectures; re-running StableLM-2 (and adding Gemma 2 and the smaller models) under the corrected hook is left to the intervention experiments described in Section~\ref{sec:limitations}. Until then we restrict the causal claim to the validated code path.
\end{document}